%% file: malar_sd_final.tex
\documentclass{easychair}

\usepackage{url}
\usepackage{xspace}
\usepackage{color}
\usepackage{array}
\usepackage{underscore}
\usepackage{tipa}
\usepackage[T1]{fontenc}
\usepackage[utf8]{inputenc}
\usepackage{rotating}

\usepackage{amsmath}
\usepackage{amssymb}
\usepackage{booktabs}
\usepackage{graphicx}
\usepackage{listings}
\usepackage{paralist}
\usepackage{subfig}
\usepackage{hyperref}

\input{thol.tex}

\newcolumntype{*}{>{\global\let\currentrowstyle\relax}}
\newcolumntype{^}{>{\currentrowstyle}}

\def\systemname#1{\textsf{#1}\xspace}

\newcommand{\MaLARea}{\systemname{MaLARea}}

\title{Machine Learner for Automated Reasoning\\ 0.4 and 0.5}
\author{Cezary Kaliszyk \inst{1} \and Josef Urban \inst{2} \and Ji\v{r}\'i Vysko\v{c}il \inst{3}}

\institute{University of Innsbruck, Austria \and
  Radboud University Nijmegen \and Czech Technical Universitity}

\titlerunning{Machine Learner for Automated Reasoning}

\authorrunning{Kaliszyk, Urban, Vysko\v{c}il}
\begin{document}
\maketitle
\begin{abstract}
  Machine Learner for Automated Reasoning (\MaLARea) is a learning and
  reasoning system for proving in large formal libraries where
  thousands of theorems are available when attacking a new
  conjecture, %
  and a large number of related problems and proofs can be used to
  learn specific theorem-proving knowledge.  The last version of the
  system has by a large margin won the 2013 CASC LTB competition. This
  paper describes the motivation behind the methods used in \MaLARea,
  discusses the general approach and the issues arising in evaluation
  of such system, and describes the Mizar@Turing100 and CASC-24
  versions of \MaLARea.

\end{abstract}

\section{Introduction: \MaLARea as an Experiment with Data-Driven
  AI/ATP Methods}

Machine Learner for Automated Reasoning (\MaLARea) is a %
(meta-)system for automated theorem proving (ATP) in large theories
consisting of thousands of formulas, symbols and proofs. The main
motivation behind the system has been to develop, employ, and evaluate
various kinds of heuristic guiding methods that are useful for
reasoning in such large (especially complex mathematical) theories. A
particular goal was to start exploring analogies with
how trained mathematicians work, i.e., how they develop and
accumulate problem-solving knowledge, learn it from others and re-use it
for attacking more and more difficult problems. Providing at least one underlying 
database for developing such methods has also been one
of the main motivations for translating the Mizar library to ATP
formats. \MaLARea started to be developed in 2007, when the
translation (MPTP~\cite{Urb04-MPTP0,Urban06}) of the whole Mizar
library to first-order logic (FOF TPTP) was largely finished. An
AI/ATP competition on a smaller set of related large-theory MPTP
problems -- the MPTP
Challenge\footnote{\url{http://www.cs.miami.edu/~tptp/MPTPChallenge/}}
-- was designed in 2006 to measure the strength of various
large-theory AI/ATP systems and techniques limited to CASC-scale
resources, and to encourage further development of such systems.
Our hope has been that the more successful AI/ATP techniques developed
experimentally for \MaLARea on the smaller benchmarks will eventually
be deployed as systems that help with mathematics done over large
formal libraries.

The system design has been influenced by earlier experiments done over
the whole translated Mizar library and its parts. The main guiding
idea tried already in~\cite{Urb04-MPTP0} was
to use the previous Mizar library proofs to learn which of the
thousands of available theorems in the library could be relevant for a
new conjecture. In~\cite{Urban06}, this was for some experiments complemented by 
learning also from the ATP proofs of related Mizar problems. 
The heuristic justification behind this overall approach to large-theory ATP was
that predicting relevance is in general very hard, i.e., it is intractable or undecidable (depending
on one's exact assumptions) to estimate how a theorem will be
  proved and what previous knowledge will best serve that. So once
such knowledge has been (often expensively) discovered for some problems, it should be
re-used as much as possible, becoming an integral part of the ATP algorithms,
instead of disregarding it and relying just on simpler (e.g., symbol-based) 
pre-programmed criteria that (at least currently) do not seem to be quite successful in emulating the 
reasoning processes leading to the human discovery of complicated mathematical proofs. 
Such an approach to algorithm design has been
recently called \emph{data-driven}\footnote{``Many of the most interesting problems in AI and computer science in
general are extremely complex often making it difficult or even
impossible to specify an explicitly programmed solution. 
Learning systems
offer an alternative methodology for tackling these problems. By
exploiting the knowledge extracted from a sample of data, they are
often capable of adapting themselves to infer a solution to such
tasks. [This approach] is referred to as the \emph{data driven}
approach, in contrast to the \emph{theory driven} approach that gives
rise to precise specifications of the required algorithms.''~\cite{KMB}} \cite{KMB}.
This paradigm does not seem yet entirely established in the domain of
Automated Reasoning (AR), 
despite its generally acknowledged crucial role %
in recent major AI (and also AR) achievements such as the IBM Watson system. %
We hope that the techniques and results described here will motivate further interest in such ATP methods.
The paper is organized as follows. Section~\ref{FirstVersions}
explains the main techniques implemented in the early versions of
\MaLARea. Section~\ref{CascLtb} briefly discusses the various
concerns and issues related to evaluating such AI/ATP
systems. Section\ref{Turing} describes the version of \MaLARea used at
the Mizar@Turing100 AI/ATP competition, and Section~\ref{CASC24} describes
the version used at the CASC-24 competition.

\section{Basic Ideas and Techniques} %
\label{FirstVersions}
The basic techniques used already by early versions of \MaLARea are
described in detail in~\cite{Urban07,US+08}.  The metasystem relies on
one or several core ATPs (originally SPASS~\cite{WeidenbachDFKSW09} and E~\cite{Sch02-AICOMM}) and one or several
machine learners (originally the naive Bayes learner provided by the
SNoW system~\cite{Carlson1999}). The goal in benchmarks like the MPTP Challenge is to
solve as many large-theory problems as possible within a global time
limit. This allows systems to spend time on analyzing and solving
different problems, and to transfer the knowledge obtained while
solving one problem to other problems. The main way how \MaLARea
does this is by incremental exploration and learning of the
\emph{relevance relation} saying which (sets of) existing theorems are
likely to be useful for proving a given conjecture. %
Making decisions about which of the problems to attack next, which
axioms to use, and how it is going to improve the overall knowledge
about the relevance relation is an instance of the general
\emph{exploration vs. exploitation} framework studied, e.g., on
multi-armed bandits~\cite{gittins1979bandit} in the domain of
reinforcement learning~\cite{sutton1998reinforcement}.

\textbf{Preprocessing, Features:} The system starts by several pre-processing steps done on the input set of all problems, during which the formulas
in the problems are syntactically normalized (using the
\texttt{tptp4X} tool from the TPTP %
distribution) and checked for
duplicate names. Different names for the same formula are merged into
one unique name. This typically later benefits the learning of the
relevance relation, which targets the formula names. More elaborate schemes for
normalization in large libraries are possible
(and could benefit the knowledge re-use even more): for example,
splitting of all conjunctions (and naming the conjuncts) and
normalizing also the symbol names using recursive content-based
hashing has been recently introduced for the whole Flyspeck
\cite{hhmcs}. The normalized formulas are then subjected to the
initial extraction of suitable features that approximate the formulas for the learning systems. In
the first version, only the set of (non-variable) symbols was used. %
In the later versions, the set of all terms in
all formulas was enumerated using a tool based on E prover's shared
term banks, and the serial numbers of the shared terms (or just their
internal string representation) were added to the set of formula
features. Overlap (measured, e.g., using the Jaccard index~\cite{jaccard01}) on such feature sets then provides a much more detailed notion of similarity between two
formulas. Since this notion of similarity can be too fine, various further
modification of this idea have been considered. One that was also
introduced early is the use of just one generic variable in all
formulas, so that terms that differ only in variables can still give
rise to similarity between formulas. Again, this idea has been in recent
related systems extended in various
ways~\cite{holyhammer,KuhlweinBKU13}. The number of such features can
be quite high (thousands to millions, depending on the size of the
data set), however for relatively simple sparse learning methods such
as naive Bayes this is not an efficiency problem when compared to the times used by the ATPs. In addition to such purely
syntactic features, much more semantic formula features were added
in~\cite{US+08} (no longer during the preprocessing phase), based on the validity of the formulas in a large pool
of finite models that are typically created dynamically as counter-examples during the main \MaLARea loop execution.
  Unlike the syntactic proximity, which will make the formulas
$\phi$ and $\neg \phi$ very close, the model-based proximity should
much better correspond to the ``true semantic relation'' of the
formulas in the Lindenbaum algebra, providing also basis for using
(and learning) straightforward (counter-)model-based criteria for
premise selection~\cite{US+08}.%

\textbf{Main Loop:} After the preprocessing steps, the system proceeds by running the main loop
which interleaves the work of the ATP systems with the work of the
learning (and other premise-selection) systems. Based on the
conjecture features and previous proofs, the learning systems try to
select the most relevant premises for the conjectures that are still
unproved. The ATP systems in turn try to (dis)prove the problems with
various numbers of the most relevant premises, adding more proofs or
counter-models to the common (initially empty) knowledge base.  There
are various options that govern this main loop, controlling mainly the
premise numbers, time limits, and the speed of re-learning on the
knowledge base when new proof (or counter-model) data become
available. The loop ends either by solving all problems, timing out,
or after trying all allowed combinations of premise numbers and time
limits for the remaining conjectures without obtaining any new
information (proof or counter-model) that could update the relevance
relation. The main purpose behind this loop is to efficiently
combine educated guessing (induction) based on the current knowledge
of the world with deductive confirmation (or disproof) of such
guesses, which produces further knowledge. The early versions of
\MaLARea outperformed other systems on (the large-theory division of) the MPTP Challenge and also on
the Mizar category of the first (2008) CASC LTB (large-theory batch)
competition.
\section{Large-Theory Competitions and Evaluating \MaLARea }
\label{CascLtb}
A major issue triggered by \MaLARea turned out to be the evaluation of
such AI systems.  Until the introduction of the LTB division in 2008
(consisting then of the Mizar, SUMO, and Cyc problems), the CASC
competition largely prohibited knowledge re-use between different
problems and systems that would come with pre-recorded information
about the solutions of other (typically TPTP) problems.\footnote{The
  fine point of such rules is of course construction of targetted ATP
  strategies and their use based on problem classification
  methods. The finer such methods are, the closer they are to
  pre-recording information about the problem solutions.}  Good
knowledge extraction and re-use is however the main research topic of
data-driven AI methods.  The MPTP Challenge addressed this dilemma by
(i) keeping the CASC rules preventing pre-recording, (ii) allowing
arbitrary recording and re-use between the problems within the competition time, (iii)
providing sufficiently many -- 252 -- related problems of varied
difficulty, and (iv) allowing further knowledge re-use by letting
systems return to unsolved problems. An even better solution (similar,
e.g., to machine learning competitions such as the Netflix Challenge)
would have been to use a public dataset for system preparation and an
unknown similar dataset for evaluation. To some extent this was
realized in the first (2008) CASC LTB competition, when the MPTP
Challenge benchmark could be used for preparing systems, but the
competition data for the Mizar category (following the same design as
the MPTP Challenge) were unknown prior to the competition.  The next
(2009) CASC LTB however significantly decreased the number of Mizar
competition problems (to 40), and focused on a query answering mode,
in which systems are additionally not allowed not to return to
unsolved problems (point (iv) above). This LTB setup was kept until
2012. A real-time query answering mode is in principle a valid
scenario in large-theory formal
mathematics~\cite{hhmcs,abs-1109-0616,KuhlweinBKU13}, but the low
number of preceding competition problems results in only a
few %
proofs to learn from, which mostly does not correspond to reality in
(formal) mathematics. Such simplification (denial of existence of
great amount of useful data) obviously largely diminishes the
usefulness of data-driven methods, whose development was the primary
motivation behind \MaLARea, MPTP, and the first large-theory
benchmarks. Consequently, \MaLARea did not compete in CASC LTB from
2009 to 2011.

\section{\MaLARea 0.4 at Mizar@Turing100}
\label{Turing}

In 2012, the Mizar@Turing100 competition brought back the possibility
of learning from many related proofs. It was based on the larger
MPTP2078 benchmark~\cite{abs-1108-3446} and followed the MPTP
Challenge rules (using 400 competition problems), modified by
additionally pre-releasing 1000 training problems together with a
large number (13455) of their (ATP or Mizar) proofs. The systems were
allowed to preprocess these problems and proofs in an arbitrary way
before the competition, and use any knowledge thus obtained in the
competition.\footnote{This was broadly motivated by the presence of a
  large number of proofs in the ITP systems' libraries, and the
  possibility to often obtain different (often shorter) ATP proofs for
  the problems exported from such libraries.} Three main additions
were made to \MaLARea, accommodating the new rules and taking into account recent large-theory research. 
\textbf{Use of the training proofs to initialize the learning systems:}
The 13455
training proofs were analyzed in the same way as the proofs obtained
during the competition (extracting the premises used), and the
resulting table became part of the competition system and used for the
first learning at the start of the competition. 
The main
interesting AI issue %
was how
to best learn the relevance relation from many different proofs (on
average 13) of the same problem. The experimentally obtained best
answer to this was (at that time, for Mizar and available learners)
to use the shortest proof
available~\cite{KuhlweinLTUH12,KuhlweinU12b}. 

\textbf{Use of non-learning predictors and their combining with learners:}
The research done
in~\cite{KuhlweinLTUH12} also motivated the second addition: using E's
version of SInE~\cite{HoderV11} to produce its own relevance ordering
of premises, which may then be also linearly combined with the
ordering produced by the machine learner. In~\cite{KuhlweinLTUH12}
such combination of rankers improved the final ATP performance by
10\%. 
\textbf{Use of ATP strategies automatically constructed on the training data:}
The newly
developed Blind Strategymaker (BliStr) strategy evolving
system~\cite{blistr} was used to automatically construct new ATP strategies on the training problems.
Building of Blind Strategymaker itself was a
response to the preliminary measurements showing that E 1.6 (used as
the only ATP in \MaLARea for the competition) could in 300s prove only
518 of the training problems, compared to 691 proved by Vampire
1.8.\footnote{Measured on pruned problems containing only the facts
  needed in the Mizar proofs.} After 6 runs (30 hours of real time on
12-core Xeon 2.67GHz server), BliStr developed a set of E strategies
that (using altogether 300s) raised the performance of E to about 650
training problems. This 25\% improvement turned out to carry over also
to the 400 competition problems, probably thanks to the relatively strong
guards against overfitting (versatility criterion) used in the BliStr
strategy-evolution loop~\cite{blistr}.  

The overall \MaLARea
parameters have then been (manually) tuned on a random subset of 100
problems from the training dataset, restricting the training data to
the remaining 900 problems. To speed up the expensive testing runs, a
file-based (content-indexed) cache of solutions was added, growing in
the end to about 2.5M unique solutions. The final \MaLARea version
solved 257 of the 400 competition problems, but lost 17 due to a bug
in proof delivery and thus came second after Vampire (which solved 248 problems). The version without the new E
strategies solved only 214 problems~\cite{blistr}.

\section{\MaLARea 0.5 at CADE-24}
\label{CASC24}

In 2013, CASC LTB (running for HOL, Isabelle and Mizar categories)
also provided pre-training problems and solutions, however they had to
be processed within the overall competition time.  The competition
problems (250 for Mizar) themselves had to be processed in the
\emph{ordered} mode, i.e, without returning to previous unsolved
problems. This made the ``explore vs. exploit'' heuristics in previous
versions of \MaLARea largely redundant, resulting in a 2013
re-implementation done from scratch which contains several new
techniques and so far omits some of the (more complicated) old ones.

\textbf{Main loop:} Instead of deciding which problem to attack next,
the system only has to decide how much time it spends in learning/predicting and
in proving for each problem. Solving previously unsolved problems might still be
beneficial for the learning~\cite{holyhammer}, but with 250
competition problems this did not seem sufficiently rewarding yet. The
main loop thus consists only of (parallelized) running of predictors,
followed by (parallelized) proving, which is (if successful) followed
by proof pseudo-minimization~\cite{holyhammer}, usually producing
better data for the learning.

\textbf{Predictors, features:} A family of distance weighted
$k$-nearest neighbor ($k$-NN) learners complemented by the IDF
(inverse document frequency) feature weighting has recently shown very
good prediction properties on the Flyspeck
dataset~\cite{EasyChair:74}. Additionally, a lot of work (measuring
distances and finding the nearest neighbors) can be simply shared in
such families for different values of $k$. The final family thus
consists of 4 subfamilies run in parallel with different IDF
weightings and feature normalization, each of them using 8 values of
$k$, producing 32 $k$-NN rankings very fast. These rankings can then
be differently sliced, easily yielding over 100 premise selections for
a given conjecture. An interesting alternative to the IDF feature
preprocessing is the latent semantic analysis (LSA), for which the
gensim toolkit\footnote{\url{http://radimrehurek.com/gensim/}} was
used, instructed to produce 400 topics (new features). Such
features were used as an input to some $k$-NNs.

\textbf{Strategies:} The BliStr system has been extended to produce
also good SInE strategies for E, and run several times on the
Mizar@Turing100 training problems and the published Flyspeck data, producing further overall strengthenings~\cite{EasyChair:74}.

\textbf{Global optimization:} Several test runs were done on
the 2012 Mizar@Turing100 competition data. The large number of premise
selections produced during these runs were collected, and used as a pool
of problems for finding the strongest combinations of the BliStr
strategies with the predictors. The final ensemble consists of 40 combinations
characterized by the selection of features, $k$-NN version, feature
weighting, premise count, and ATP strategy.

\textbf{Evaluation and Competition Performance:}
The final system solved
260 of the 400 Mizar@Turing100 competition problems (used for optimization here). On the unknown CASC-24 LTB competition data the system  solved 239
problems out of 750. The second best system (E
1.8-LTB) solved 135 problems. 
This is the largest relative distance
(77\% more) between the first and second system in CASC since
Waldmeister's victory in the UEQ division in 2000.

\section{Acknowledgements}

It has been a great pleasure to use Stephan Schulz's open-source E prover.
We now use E as the main ground ATP, E's
implementation of SInE, its API for strategy specification, and its code base
for our fast shared-term enumerator.

\begin{small}
\bibliographystyle{abbrv}
\bibliography{ate11}
\end{small}
\end{document}

%% file: thol.tex
\makeatletter

\def\holdocspecials{\do\ \do\$\do\&%
  \do\#\do\^\do\^^K\do\_\do\^^A\do\%}

\def\holtt{\trivlist \item[]\if@minipage\else\vskip\parskip\fi
\leftskip\@totalleftmargin\rightskip\z@
\parindent\z@\parfillskip\@flushglue\parskip\z@
\@tempswafalse \def\par{\if@tempswa\hbox{}\fi\@tempswatrue\@@par}
\obeylines \tt \let\do\@makeother \holdocspecials
 \frenchspacing\@vobeyspaces}

\makeatother

\newlength{\hsbw}
\newcommand\HOLSpacing{13pt}

   \newcommand\hilbert{\varepsilon}

   \newcommand{\Cond}{\(\rightarrow\)}
   \newcommand{\Eqv}{\(\equiv\)}
   \newcommand{\Iff}{\(\Longleftrightarrow\)\hspace{-1.5mm}}
   \newcommand{\Fa}{\(\forall\)}
   \newcommand{\Et}{\(\exists\)}
   \newcommand{\Eu}{\(\exists_{unique}\)}

   \newcommand{\Impl}{\(\Longrightarrow\)\hspace{-1.5mm}}
   \newcommand{\Func}{\(\to\)\hspace{-1.5mm}}

   \newcommand{\Lam}{\(\lambda\)}
   
   \newcommand{\Minus}{\(-\)}
   \newcommand{\Lminus}{\(-\)\hspace{-1.5mm}}
   \newcommand{\Prime}{\('\)}
   \newcommand{\Und}{\_}
   \newcommand{\Lt}{\(<\)}
   \newcommand{\Gt}{\(>\)}
   \newcommand{\Leq}{\(\leq\)}
   \newcommand{\Geq}{\(\geq\)}
   \newcommand{\Eq}{\(=\)}
   \newcommand{\Lrb}{\((\)}
   \newcommand{\Rrb}{\()\)}

   \newcommand{\Next}{\(\bigcirc\)}
   \newcommand{\Prev}{\(\ominus\)}
   \newcommand{\WPrev}{\(\widetilde{\bigcirc}\)}
   \newcommand{\Event}{\(\Diamond\)}
   \newcommand{\Once}{\(\underline{\Diamond}\)}  
\newcommand{\Hilbert}{\(\hilbert\)}

\newcommand{\Conj}{\(\wedge\)}
\newcommand{\Disj}{\(\vee\)}
\newcommand{\Neg}{\(\neg\)}
\newcommand{\Pnd}{\(\Diamond\)}

\newcommand{\Models}{\(\models\)}

\long\def\rechol#1#2#3{\let\next=\rechol\def\postnext{#2#3}\ifx#1\end
\let\next=\relax\def\postnext{\relax}
\else\ifx#1!\Fa                                          
\else\ifx#1@\Hilbert                                     
\else\ifx#1\#\Pnd                                        
\else\ifx#1'\Prime                                       
\else\ifx#1~\Neg                                         
\else\ifx#1\~\Neg
\else\ifx#1_\Und                                         
\else\ifx#1(\ifx#2+\ifx#3)\Next\def\postnext{}\fi        
            \else\ifx#2-\Prev\def\postnext{}             
            \else\ifx#2~\ifx#3)\WPrev\def\postnext{}\fi            
             \else\Lrb\fi\fi\fi                          
\else\ifx#1)\Rrb%
\else\ifx#1\/\Disj                                       
\else\ifx#1\.\Lam                                        
\else\ifx#1>\ifx#2=\Geq\def\postnext{#3}\else\Gt\fi      
\else\ifx#1?\ifx#2!\Eu\def\postnext{#3}\else\Et\fi       
\else\ifx#1-\ifx#2>\Func\def\postnext{#3}               
            \else\ifx#2-\Lminus\def\postnext{#3}
            \else\Minus\fi\fi                               
\else\ifx#1|\ifx#2-\Turns\def\postnext{#3}               
            \else\ifx#2=\Models\def\postnext{#3}
                 \else\Bar\fi\fi
\else\ifx#1<\ifx#2=\ifx#3>\Iff\def\postnext{}       
                   \else\Leq\def\postnext{#3}\fi    
            \else\ifx#2+\Event\def\postnext{}       
            \else\ifx#2-\Once\def\postnext{}       
            \else\Lt\fi\fi\fi                       
\else\ifx#1=\ifx#2=\ifx#3>\Impl\def\postnext{}            
                   \else\Eqv\def\postnext{#3}\fi         
            \else\ifx#2>\Cond\def\postnext{#3}
                 \else\Eq\fi\fi
\else\ifx#1/\ifx#2\^^M\Conj\par\def\postnext{#3}         
            \else\ifx#2\ \Conj\ \def\postnext{#3}\else#1\fi\fi  
\else#1\fi\fi\fi\fi\fi\fi\fi\fi\fi\fi\fi\fi\fi\fi\fi\fi\fi\fi\fi
\expandafter\next\postnext}